\documentclass{llncs}

\usepackage{graphicx}
\usepackage{amsmath}
\usepackage{amssymb}
\usepackage{subfig}

\usepackage{algorithm}
\usepackage{algorithmicx}
\usepackage[noend]{algpseudocode}

\usepackage{pgfplots}

\pgfplotsset{vasymptote/.style={
    before end axis/.append code={
        \draw[densely dashed] ({rel axis cs:0,0} -| {axis cs:#1,0})
        -- ({rel axis cs:0,1} -| {axis cs:#1,0});
    }
}}

\usepackage{mathtools}

\usepackage{verbatim}

\newcommand{\stateSort}{\mathcal{S}}

\newcommand{\actionSort}{\mathcal{A}}

\newcommand{\nodeSort}{\mathcal{V}}

\newcommand{\programSort}{\mathcal{P}}

\newcommand{\querySort}{\mathcal{Q}}

\newcommand{\argmax}{\operatornamewithlimits{argmax}}

\title{Monte Carlo Action Programming}
\author{Lenz Belzner}
\institute{Institute for Informatics\\LMU Munich}
	
\begin{document}

\maketitle

\begin{abstract}
This paper proposes Monte Carlo Action Programming, a programming language framework for autonomous systems that act in large probabilistic state spaces with high branching factors.
It comprises formal syntax and semantics of a nondeterministic action programming language. The language is interpreted stochastically via Monte Carlo Tree Search.
Effectiveness of the approach is shown empirically.
\end{abstract}


\section{Introduction}

We consider the problem of sequential decision making in highly complex and changing domains. These domains are characterized by large probabilistic state spaces and high branching factors. Additional challenges for system design are occurrence of unexpected events and/or changing goals at runtime.

A state of the art candidate for responding to this challenge is behavior synthesis with online planning \cite{browne2012survey,keller-helmert-icaps2013,kolobov2012reverse}. Here, a planning agent evaluates possible behavioral choices w.r.t. current situation and background knowledge at runtime. At some point, it acts according to this evaluation and observes the actual outcome of the action. Planning continues, incorporating the observed result. Planning performance directly correlates with search space cardinality. 

This paper introduces \textit{Monte Carlo Action Programming} (MCAP) to reduce search space cardinality through specification of heuristic knowledge in the form of procedural nondeterministic programs. MCAP is based on stochastic interpretation of nondeterministic action programs by Monte Carlo Tree Search (MCTS) \cite{browne2012survey,chaslot2006monte}. Combining search space constraints and stochastic interpretation enables program evaluation in large probabilistic domains with high branching factors.
From the perspective of online planning, MCAP provides a formal non-deterministic action programming language that allows to specify plan sketches for autonomous systems. From the perspective of action programming, MCAP introduces stochastic interpretation with MCTS. This enables effective program interpretation in very large, complex domains.

We will discuss MCTS and action programming in Section \ref{sec:preliminaries}. Section \ref{sec:MCAP} introduces MCAP. In Section \ref{sec:evaluation} we empirically compare MCTS and MCAP specifications for online planning. We conclude and sketch venues for further research in Section \ref{sec:conclusion}.

\section{Related Work}
\label{sec:preliminaries}

We briefly review Monte Carlo Tree Search in Section \ref{sec:MCTS} and action programming in Section \ref{sec:actionprogramming}.

\subsection{Monte Carlo Tree Search}
\label{sec:MCTS}

Monte Carlo Tree Search (MCTS) is a framework for statistical search in very large state spaces with high branching factors based on a generative model of the domain (i.e. a simulation). It yields good performance even without heuristic assessment of intermediate states in the search space. The MCTS framework originated from research in computer Go \cite{chaslot2006monte,DBLP:conf/aips/SilverSM13}. The game Go exposes the mentioned characteristics. Also, not many good heuristics are known for Go. Nevertheless, specialized Go programs based on the MCTS algorithm are able to play on the niveau of a human professional player \cite{DBLP:journals/cacm/GellyKSSSST12}. MCTS is also commonly used in autonomous planning \cite{keller-helmert-icaps2013,kolobov2012reverse} and has been applied successfully to a huge number of other search tasks \cite{browne2012survey}.

\begin{figure}
	\centering
	\includegraphics[width=\columnwidth]{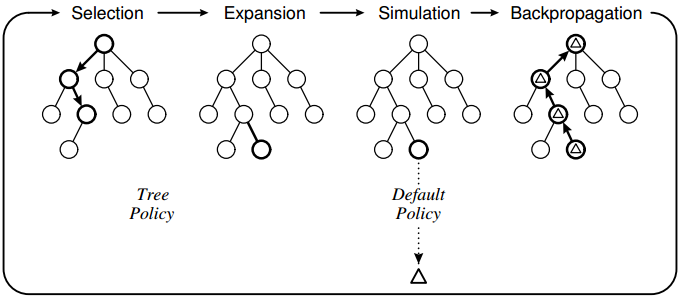}
	\caption{Monte Carlo Tree Search \cite{browne2012survey}.}
	\label{fig:MCTS}
\end{figure}

MCTS adds nodes to the tree iteratively. Nodes represent states and store metadata about search paths that lead through them. Gathered metadata comprises mean reward (i.e. node value) and the number of searches that passed through the node. It enables assessment of exploration vs. exploitation: Should search be directed to already explored, promising parts of the search space? Or should it gather information about previously unexplored areas?

Figure \ref{fig:MCTS} shows the basic principle of MCTS. Based on node information, MCTS selects an action w.r.t. a given \textit{tree policy}. The successor state is determined by simulating action execution. Selection is repeated as long as simulation leads to a state that is represented by a node in the tree. Otherwise, a new node representing the simulated outcome state is added to the tree (expansion). Then, a \textit{default policy} is executed (e.g. uniform random action execution). Gathered reward is stored in the new node (simulation or \textit{rollout}). This gives a first estimation of the new node's value. Finally, the rollout's value is backpropagated through the tree and the corresponding node values are updated. MCTS repeats this procedure iteratively.
Algorithm \ref{alg:mcts} shows the general MCTS approach in pseudocode. Here, $v_0$ is the root node of the search tree. $v_l$ denotes the last node visited by the tree policy. $\Delta$ is the value of the rollout from $v_l$ according to the default policy.
\begin{algorithm}
	\begin{algorithmic}[1]
		\Procedure {mcts}{$s_0$}
		\State create root node $v_0$ with state $s_0$
		\While {within computational budget}
		\State $v_l \gets$ \Call{treepolicy}{$v_0$}
		\State $\Delta \gets$ \Call{defaultpolicy}{$v_l$}
		\State \Call{backup}{$v_l,\Delta$}
		\EndWhile
		\State \Return a(\Call{bestchild{$v_0$}})
		\EndProcedure
	\end{algorithmic}
	\caption{General MCTS approach \cite{browne2012survey}}
	\label{alg:mcts}
\end{algorithm}

MCTS can be interrupted at any time and yields an estimation of quality for all actions in the root state. The best action (w.r.t. node information) is executed and its real outcome is observed. MCTS continues reusing the tree built so far. Eventually, nodes representing past states are pruned from the tree.

\subsection{Action Programming}
\label{sec:actionprogramming}

Nondeterministic action programs define sketches for system behavior that are interpreted at runtime, leaving well-defined choices to be made by the system at runtime. Interpreting an action program typically provides a measure of quality for particular instantiations of theses sketches. Concrete traces are then executed w.r.t. to this quality metric.

Well-established action programming languages are Golog \cite{bib:de2000congolog,bib:degiacomo2009} and Flux \cite{bib:thielscher2005}. Each is interpreted w.r.t. a particular formal specification of domain dynamics: The situation calculus and the fluent calculus are concerned with specification of action effects and domain dynamics in first order logic \cite{bib:boutilier2000decision,bib:thielscher1998}. For both Golog and Flux, Prolog interpreters have been implemented.


The MCAP framework differs from these formalisms and their respective languages: (a) MCAP does not provide nor require a specific formal representation of domain dynamics. Rather, any form of domain simulation suffices. (b) MCAP does not explore the search space exhaustively. Rather, programs are interpreted stochastically by MCTS. The search space is explored iteratively. Program interpretation is directed to promising areas of the search space based on previous interpretations. Search can be interrupted any time yielding an action recommendation accounting for current situation and a given program. Recommendation quality depends on the number of simulations used for search \cite{browne2012survey}.

\section{Monte Carlo Action Programming}
\label{sec:MCAP}

This Section introduces Monte Carlo Action Programming (MCAP), a nondeterministic procedural programming framework for autonomous systems. 
The main idea of the MCAP framework is to allow to specify behavioral blueprints that leave choices to an agent. A MCAP is a nondeterministic program. MCAP programs are interpreted probabilistically by MCTS. MCAPs constrain the MCTS search space w.r.t. a procedural nondeterministic program. 


\subsection{Framework Parameters}
\label{sec:parameters}

The MCAP framework requires the following specification.

\begin{enumerate}
	\item A \textit{generative domain model} that captures the probability distribution of successor states w.r.t. current state and executed action (Equation \ref{eq:simulate}). The model does not have to be explicit: The framework only requires a simulation that allows to query one particular successor state.
	\begin{align}
	\label{eq:simulate}
	\mathrm{simulate} : P(\stateSort ~ | ~ \stateSort \times \actionSort)\\
	\end{align}
	\item A reward function $R$ that encodes the quality of a state w.r.t. system goals (Equation \ref{eq:reward}).
	\begin{align}
	\label{eq:reward}
	\mathcal{R} &: \stateSort \rightarrow \mathbb{R}
	\end{align}
	\item A discount factor $\gamma \in [0;1]$ weights the impact of potential future decision on the current situation. A discount factor of zero means that only immediate consequences of action are considered. A discount factor of one means that all future consequences influence the current decision equally, regardless of their temporal distance.
	\item A maximum search depth $h_\mathrm{max} \in \mathbb{N}$.
\end{enumerate}

\subsection{Syntax}
\label{sec:syntax}

Equation \ref{eq:syntax} defines syntax of the MCAP language. $\epsilon$ is the empty program, $\actionSort$ denotes specified action space, $\texttt{;}$ is a sequential operator, $+$ is nondeterministic choice, $\parallel$ denotes interleaving concurrency. $Q$ denotes the query space for conditional evaluation (see Equation \ref{eq:query}). $?$ denotes querying the current execution context. $\circ$ denotes a conditional loop.
\begin{align}
\label{eq:syntax}
\programSort := \epsilon ~&\vline~ \actionSort ~\vline~ \programSort \texttt{;} \programSort 
~\vline~ \programSort \texttt{+} \programSort ~\vline~ \programSort \parallel \programSort \nonumber \\
~ &\vline ~ ?(\querySort) \{ \programSort \} ~ \vline ~ \neg ?(\querySort) \{ \programSort \} 
~ \vline ~ \circ \{ \querySort \} \{ \programSort \}
\end{align}

\textbf{Normal Form}~
We define a normal form $\programSort_\mathrm{norm}$ for MCAPs. Each program in normal form is a choice of programs with an action prefix and any tail program.
\begin{align}
	\programSort_\mathrm{norm} := \sum \left( \actionSort \texttt{;} \programSort \right)
\end{align}

Equations \ref{eq:normalFormBegin} to \ref{eq:normalFormEnd} define a term reduction system that ensures transformation of programs to their normal form.
\begin{align}
\label{eq:normalFormBegin}
\epsilon \texttt{;} p &= p\\
p + p &= p \\
(p_1 + p_2) \texttt{;} p &= (p_1 \texttt{;} p) + (p_2 \texttt{;} p)\\
p \texttt{;} (p_1 + p_2) &= (p \texttt{;} p_1) + (p \texttt{;} p_2)\\
p_1 \parallel (p_2 + p_3) & = (p_1 \parallel p_2) + (p_1 \parallel p_3) \\
\left( a_1 \texttt{;} p_1 \right) \parallel \left( a_2 \texttt{;} p_2 \right) & = 
\left( a_1 \texttt{;} (p_1 \parallel \left( a_2 \texttt{;} p_2 \right) ) \right)
\nonumber\\
& \hspace{1em} + 
\left( a_2 \texttt{;} (\left( a_1 \texttt{;} p_1 \right) \parallel p_2) \right) \\
a_1 \parallel (a_2 \texttt{;} p) & = (a_1 \texttt{;} a_2 \texttt{;} p) + (a_2 \texttt{;} (a_1 \parallel p)) \\
a_1 \parallel a_2 & = (a_1 \texttt{;} a_2) + (a_2 \texttt{;} a_1)
\label{eq:normalFormEnd}
\end{align}

\subsection{Semantics}
\label{sec:semantics}

This Section formalizes MCAP semantics in the context of MCTS interpretation.

\textbf{Search Tree}~
We introduce a formal representation of the search tree. Its purpose is to accumulate information about computation traces w.r.t. simulation and system action choices. Tree nodes represent states $\in \stateSort$ and actions $\in \actionSort$. State nodes $\nodeSort_\stateSort$ and action nodes $\nodeSort_\actionSort$ alternate (Equations \ref{eq:stateNode} and \ref{eq:actionNode}). Nodes contain aggregation of metadata $\mathcal{D}$ that guides further search. Aggregated data are visitation count and node value (Equation \ref{eq:metadata}).
\begin{align}
\nodeSort_\stateSort &\subseteq \stateSort \times \mathcal{D} \times 2^{\nodeSort_\actionSort} \label{eq:stateNode}\\
\nodeSort_\actionSort &\subseteq \actionSort \times \mathcal{D} \times 2^{\nodeSort_\stateSort} \times \programSort \label{eq:actionNode}\\
\mathcal{D} &\subseteq \mathbb{N} \times \mathbb{R} \label{eq:metadata}
\end{align}

While it is possible to use a DAG instead of a tree \cite{saffidine2012ucd}, we will concentrate on the tree setting in this paper for the sake of simplicity.

\textbf{Framework Operations}~
Equations \ref{eq:signaturesBegin} to \ref{eq:signaturesEnd} show the functional signatures of MCAP framework operations. We will define each one in the rest of this Section. 
\begin{align}
\label{eq:signaturesBegin}
\mathrm{select} &: \nodeSort_\stateSort \rightarrow \nodeSort_\actionSort\\
\mathrm{expand} &: \stateSort \times \programSort \rightarrow \nodeSort_\stateSort\\
\mathrm{rollout} &: \stateSort \times \programSort \times \mathbb{N} \rightarrow \mathbb{R}\\
\mathrm{update} &: \nodeSort_\stateSort \rightarrow \nodeSort_\stateSort\\
\mathrm{update} &: \nodeSort_\actionSort \rightarrow \nodeSort_\actionSort
\label{eq:signaturesEnd}
\end{align}

\textbf{Selection}~
Equation \ref{eq:select} shows UCB1 action selection. It is a popular instantiation of the MCTS tree policy based on regret minimization \cite{kocsis2006bandit,auer2002finite}. $q(v_a)$ denotes the current value aggregated in the metadata of action node $v_a$. $\texttt{\#}(v_s)$ and $\texttt{\#}(v_a)$ denote the number of searches that visited the corresponding node stored in its metadata (see also Algorithm \ref{alg:mcap}, lines 2 and 10). UCB1 favors actions that expose high value (first term of the sum), and adds a bias towards actions that have not been well explored (second term of the sum). The parameter $c$ is a constant to control the tendency towards exploration.
\begin{align}
\label{eq:select}
\mathrm{select}(v_s) &= \mathrm{argmax}_{v_a \in \vec{v_a}(v_s)} \left( q(v_a) + c \cdot \sqrt{\dfrac{2 \ln{\texttt{\#}(v_s)}}{\texttt{\#}(v_a)}} \right) 
\end{align}

\textbf{Queries}~
Our framework requires specification of a query representation and a satisfaction function of queries and states to enable conditional computation. Queries $Q$ are evaluated w.r.t. a given state $\in \stateSort$ and yield a set of substitutions for query variables (Equation \ref{eq:query}). It returns the set of substitutions for variables in the query for which the query holds in the state. In case the query is ground and holds, the set containing the empty substitution $\{ \emptyset \}$ is returned. If the query does not hold, it returns the empty set $\emptyset$. We write $\vdash$ in infix notation and $s \not\vdash q \Leftrightarrow s \vdash q = \emptyset$.
\begin{align}
\label{eq:query}
\vdash : \querySort \times \stateSort \rightarrow 2^\Theta
\end{align}

\textbf{Interpretation of MCAPs}~
Expansion of the tree is constrained by a given MCAP through interpreting it w.r.t a given state. The \textit{potential program function} constrains the search space w.r.t. given action program and current system state. It maps an MCAP and a given state to the set of normalized MCAPs that result from (a) nondeterministic choices and (b) interpretations of queries.
\begin{align}
\mathrm{pot} : \stateSort \times \programSort \rightarrow 2^{\programSort_\mathrm{norm}}
\end{align}

Equations \ref{eq:potBegin} to \ref{eq:potEnd} define MCAP interpretation by the potential program function inductively on the structure of $\programSort$.
\begin{align}
\label{eq:potBegin}
\mathrm{pot}\left(s, \epsilon \right) &= \emptyset\\
\mathrm{pot}\left(s, a \right) &= \{ a \texttt{;} \epsilon \}\\
\mathrm{pot} \left( s, p \texttt{;} p' \right) &= \bigcup_{p'' \in \mathrm{pot}(s, p)} \left( p'' \texttt{;} p' \right)\\
\mathrm{pot}\left(s, \sum_i p_i \right) &= \bigcup_i \mathrm{pot}(s, p_i)\\
\mathrm{pot}(s, ?\{ q \}\{ p \}) &= \bigcup_{\theta \in s \vdash q} \mathrm{pot}\left(s, \theta(p) \right)\\
\mathrm{pot}(s, \neg?\{ q \}\{ p \}) &=
\begin{cases}
\mathrm{pot}(s, p) &\mbox{\textbf{ if }} s \not\vdash q\\
\emptyset &\mbox{\textbf{ otherwise}}
\end{cases}\\
\mathrm{pot}(s, \circ\{ q \}\{ p \}) &= \mathrm{pot}(s, ?\{ q \}\{ p \} \texttt{;} \circ\{ q \}\{ p \})
\label{eq:potEnd}
\end{align}

\textbf{Expansion}~
Equation \ref{eq:expansion} shows the MCAP expansion mechanism. $s \in \stateSort$ denotes the state for which a new node is added. $p$ is the MCAP to be executed in state $s$. Potential programs $pot(s,p)$ in normal form define the set of action node children for actions $a$ that contain the corresponding tail programs $p'$. Thus, a MCAP effectively constrains the search space. $d_0 \in \mathcal{D}, d_0 = (0,0)$ defines initial node metadata.
\begin{align}
\label{eq:expansion}
	\mathrm{expand}(s,p) = (s,d_0,\vec{v_a}) & \nonumber \\
	\text{where} ~ \vec{v_a} &= \bigcup_{(a,p') \in \mathrm{pot}(s,p)} (a,d_0,\emptyset,p')
\end{align}

\textbf{Rollout}~
After expansion a rollout is performed. A number of simulation steps is performed (i.e. until maximum search depth $h_\mathrm{max}$ is reached) and the reward for resulting states is aggregated. An MCAP $p$ defines the rollout's default policy. Actions and corresponding tail programs are selected uniformly random from the set of potential programs in each state $s$ encountered in the rollout.
\begin{align}
	&\mathrm{rollout}(s, p, h) = \nonumber \\
	& \hspace{2em}
	\begin{cases}
	R(s) & \mbox{\textbf{ if }} h = h_\mathrm{max} \\
	R(s) + \gamma \cdot \mathrm{rollout}(s', p', h + 1) &\mbox{\textbf{ otherwise}}
	\end{cases} \nonumber \\
	&  \hspace{2em} \text{where} ~ (a,p') \sim \mathrm{pot}(s,p) \wedge s' \sim \mathrm{simulate}(s' | s, a)
\end{align}

\textbf{Value Update}~
After a node is expanded its value is determined by a rollout. The newly created value is then incorporated to the search tree by value backpropagation along the search path. In general any kind of value update mechanism is feasible, e.g. a mean update as used by many MCTS variants. MCAP uses dynamic programming (i.e. a Bellman update) for updating node values \cite{bib:bellmanDynamicProgramming}. An action's value is the weighted sum of its successor states' values (Equation \ref{eq:actionValue}). A state's value is the currently obtained reward and the value of the currently optimal action (Equation \ref{eq:stateValue}).
\begin{align}
\label{eq:actionValue}
\mathrm{update}(v_a) &= \sum_{v_s \in \vec{v_s}(v_a)} \dfrac{\texttt{\#}(v_s)}{\texttt{\#}(v_a)} v(v_s)\\
\label{eq:stateValue}
\mathrm{update}(v_s) &= R(s(v_s)) + \max_{v_a \in \vec{v_a}(v_s)} q(v_a)
\end{align}

Algorithm \ref{alg:mcap} shows the interplay of selection, aggregation of metadata, simulation, expansion, rollout and value update for Monte Carlo Action Programming.
\begin{algorithm}
	\begin{algorithmic}[1]
	\vspace{0.5em}
	\Require{$h_\mathrm{max}, R, \mathrm{pot}, \mathrm{simulate}$}
	\vspace{0.5em}
	
	\Procedure{mcap}{$v_s, h$}
	
	\State $\texttt{\#}(v_s) \gets \texttt{\#}(v_s) + 1$ \Comment{increase state node count}
	
	\If {$h = h_\mathrm{max}$}
	\State \Return \Comment{reached maximum search depth}
	\EndIf
	
	\If {$\vec{v_a}(v_s) = \emptyset$}
	\State \Return \Comment{no action is available}
	\EndIf
	
	\State $v_a \gets \mathrm{select}(v_s)$ \Comment{select action node}
	\State $\texttt{\#}(v_a) \gets \texttt{\#}(v_a) + 1$ \Comment{increase action node count}
	\State $s' \sim \mathrm{simulate}(v_a)$ \Comment{simulate action outcome}
	\If {$\exists v_{s'} \in \vec{v_s}(v_a) : s(v_{s'}) = s'$} \Comment{successor exists}
	\State \Call{mcap}{$v_{s'}, h + 1$} \Comment{recursive call through the tree}
	\State $v_a \gets \mathrm{update(v_a)}$ \Comment{update action quality} 
	\State $v_s \gets \mathrm{update(v_s)}$ \Comment{update state value} 
	\Else
	\State $v_{s'} \gets \mathrm{expand}(s', p(v_a))$ \Comment{create successor node}
	\State $r \leftarrow \mathrm{rollout}(s', p(v_a), h)$ \Comment{estimate node value}
	\State $d(v_{s'}) \gets (0, r)$ \Comment{set state node metadata}
	\State $\vec{v_s}(v_a) \leftarrow \vec{v_s}(v_a) \cup \{ v_{s'} \}$
	\Comment{add successor node}
	\EndIf
	\EndProcedure
	\end{algorithmic}
	\caption{Monte Carlo Action Programming}
	\label{alg:mcap}
\end{algorithm}

Algorithm \ref{alg:mcapOnline} shows the integration of MCAP with online planning. While the system is running, a given MCAP is repeatedly evaluated and executed until termination (lines 2 -- 4).
Evaluation is performed by MCTS until a certain budget is reached (lines 6 -- 8). The currently best action w.r.t. MCAP interpretation is determined (line 9). If there is no such action, the program terminated (line 10). Otherwise, the best action is executed and the outcome observed (lines 13 and 14). In case the new state is already represented in the search tree, the corresponding state node is used as new root for further search (lines 15 and 16). Otherwise, a new root node is created (line 18).
\begin{algorithm}
	\begin{algorithmic}[1]
		\vspace{0.5em}
		\Require{initial state $s_\mathrm{init}$, MCAP $p_\mathrm{init}$, budget}
		\vspace{0.5em}
		
		\State $v_\mathrm{init} \leftarrow \mathrm{expand}(s_\mathrm{init},p_\mathrm{init})$
		\Comment {initial node}
		\While {running}
		\State \Call{online-mcap}{$v_\mathrm{init}$}
		\EndWhile
		
		\vspace{0.5em}
		
		\Procedure{online-mcap}{$v_s$}
		\While {budget}
		\State \Call{mcap}{$v_s$}
		\Comment {repeatedly update $v_s$ w.r.t. MCAP}
		\EndWhile
		\State $v_a^{\mathrm{max}} \gets \argmax_{v_a \in \vec{v_a}(v_s)} q(v_a)$
		\Comment {get optimal action}
		
		\If {$v_a^{\mathrm{max}} = \mathrm{null}$}
			\State \Return \Comment{MCAP terminated}
		\EndIf
		
		\State execute $a(v_a^{\mathrm{max}})$
		\Comment {execute action}
		\State observe $s'$
		\Comment {observe the outcome}
		\If {$v_{s'} \in \vec{v_s}(v_a^{\mathrm{max}})$}
		\Comment{previously considered}
			\State $v_s \gets v_{s'}$ \Comment{reuse planning result}
		\Else
		\Comment{previously unconsidered}
			\State $v_s \gets \mathrm{expand}(v_s, p(v_a^{\mathrm{max}}))$
		\EndIf
		
		\EndProcedure
	\end{algorithmic}
	\caption{Online MCAP}
	\label{alg:mcapOnline}
\end{algorithm}

\section{Experimental Evaluation}
\label{sec:evaluation}

\subsection{Example Domain}
\label{sec:exampleDomain}

\begin{figure}
\centering
\includegraphics[width=\columnwidth]{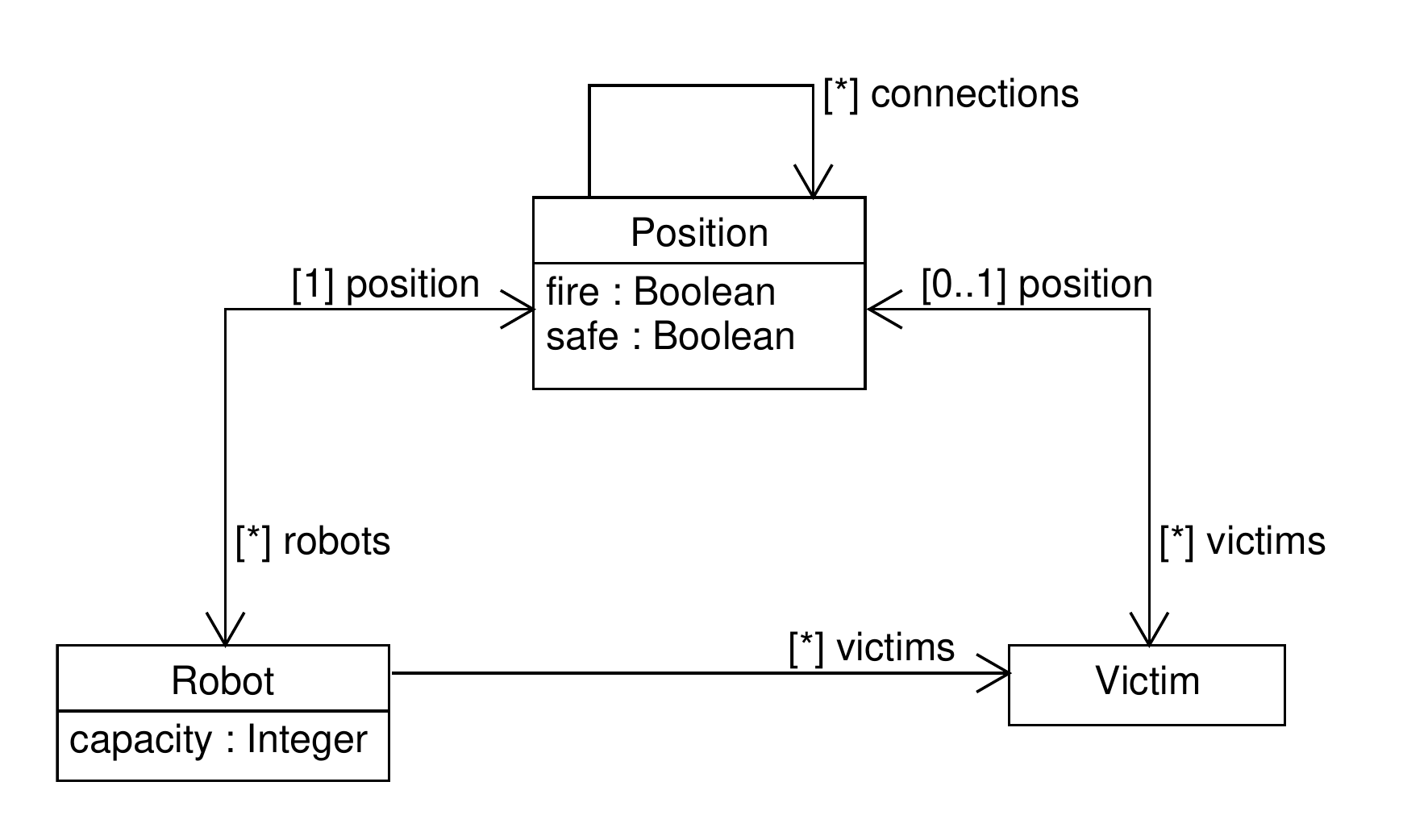}
\caption{
Class diagram of the rescue domain.
}
\label{fig:rescueClassDiagram}
\end{figure}

We introduce the \textit{rescue domain} as illustrating example.
Robots can move around a connected graph of positions and lift or drop victims.
The number of victims a robot can carry is limited by its capacity.
A position may be on fire, in which case a robot cannot move there. 
At every time step the fire attribute of a position may change
 depending on how many of the position's neighbors are on fire.
A \textit{safe} position never catches fire.
The class diagram of the rescue domain is shown in Figure \ref{fig:rescueClassDiagram}.
A particular state of the domain is an instantiation of this class diagram.

Possible system actions are:
\begin{enumerate}
\item
$\textit{Move}(R,P)$:
Robot $R$ moves to target position $P$ if it is connected to the robot's current position and is not on fire.
\item
$\textit{Extinguish}(R,P)$:
Robot $R$ extinguishes fire at a neighbor position $P$.
\item
$\textit{Lift}(R,V)$:
Robot $R$ lifts victim $V$ (at same location) if it has capacity left.
\item
$\textit{Drop}(R,V)$: Robot $R$ drops lifted victim $V$ at the current location.
\item
$\textit{Noop}$: Does nothing.
\end{enumerate}

\subsection{Setup \& Results}

Effectiveness of the MCAP framework was evaluated empirically for the rescue domain. A simulation of the domain  was used as generative model. Reward $R(s)$ was defined as the number of victims located at safe positions in state $s$. Also, each victim not burning provided a reward of 0.1. Maximum search depth was set to $h_\mathrm{max} = 40$ and the discount factor was set to $\gamma = 0.9$.
Experiments were conducted with randomized initial states, each consisting of twenty positions with 30\% connectivity. Three positions were safe, ten victims and ten fires were located randomly on unsafe positions. Robot capacity was set to two. This setup yields a state space containing more than $10^{19}$ possible states.
Fires ignited or ceased probabilistically at unsafe positions. Actions succeeded or failed probabilistically (p = 0.05). This yields a branching factor of $2 \cdot 2^{17}$ for each action.

In the experiments using plain MCTS all actions $\in \actionSort$ were evaluated at each step. Algorithm \ref{alg:mcapExperiment} shows pseudocode for the program used to determine the action to evaluate in the experiments with MCAP. Both MCTS and MCAP used 1000 playouts at each step for action evaluation.
\begin{algorithm}
	\begin{algorithmic}[1]
		\While {true}
		\If {$\mathrm{self.position.safe} \wedge \mathrm{self.victims} \neq \emptyset$}
		\State $\sum_{v \in \mathrm{self.victims}} \mathrm{self.drop}(v)$
		\ElsIf {$\neg(\mathrm{self.pos.safe}) \wedge \mathrm{self.position.victims} \neq \emptyset$}
		\State $\sum_{v \in \mathrm{self.position.victims}} \mathrm{self.lift}(v)$
		\Else
		\State $\sum_{a \in \actionSort} a$
		\EndIf
		\EndWhile
	\end{algorithmic}
	\caption{Pseudocode of the MCAP used in the experiments}
	\label{alg:mcapExperiment}
\end{algorithm}

System performance was measured with the statistical model checker Multivesta \cite{sebastio2013multivesta}. Two metrics of system behavior with and without MCAP search space constraints were assessed: Ratios of safe victims and burning victims.

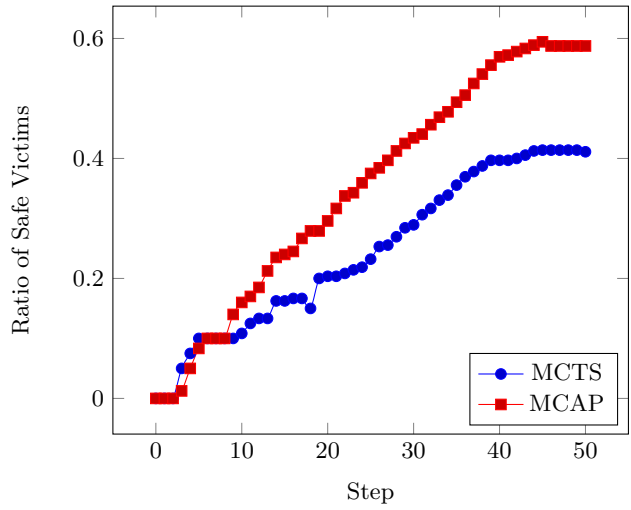
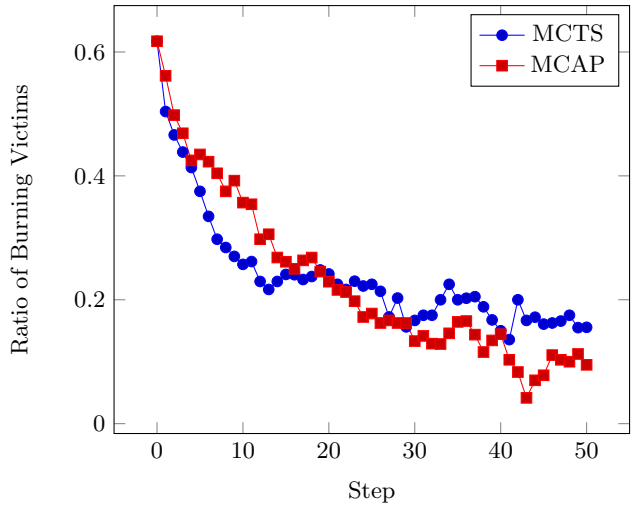
\begin{figure}
	\centering
	\subfloat[]{
		\begin{tikzpicture}[scale=1]
		\begin{axis}[
		xlabel=Step,
		ylabel=Ratio of Safe Victims,
		mark options={scale=1},
				smooth,
				legend style ={ at={(0.98,0.02)}, 
					anchor=south east, draw=black, 
					fill=white,align=left}
		]
		
		\addplot+ 
		table [x index=0, y index=1, y error index=2, col sep=space] {experiments/mctsBase.txt};
		\addlegendentry{MCTS}
		\addplot+ 
		table [x index=0, y index=1, y error index=2, col sep=space] {experiments/mcapBase.txt};
		\addlegendentry{MCAP}
		\end{axis}
		\end{tikzpicture}
		\label{fig:resultsBaseSafe}
	}
	\vfill
	\subfloat[]{
		\begin{tikzpicture}[scale=1]
		\begin{axis}[
		xlabel=Step,
		ylabel=Ratio of Burning Victims,
		mark options={scale=1},
		]
		
		\addplot+ 
		table [x index=0, y index=3, y error index=4, col sep=space] {experiments/mctsBase.txt};
		\addlegendentry{MCTS}
		\addplot+ 
		table [x index=0, y index=3, y error index=4, col sep=space] {experiments/mcapBase.txt};
		\addlegendentry{MCAP}
		\end{axis}
		\end{tikzpicture}
		\label{fig:resultsBaseBurning}
	}
	
	\caption{Comparison of (a) safe victims and (b) burning ratios for MCTS and MCAP.}
	\label{fig:resultsBase}
\end{figure}

Figure \ref{fig:resultsBase} compares the average results for behavior synthesis with plain MCTS and with MCAP within a 0.1 confidence interval. The effect of MCAP search space reduction on system performance can clearly be seen. The configuration making use of online MCAP interpretation achieves larger ratios of safe victims and manages the reduction of burning victim ratios better than the configuration not making use of MCAP. With plain MCTS, search is distracted by low reward regions due to avoiding burning victims. MCAP search identifies high reward regions where victims are saved within the given budget. 

\begin{figure}
	\centering
	\subfloat[]{
		\begin{tikzpicture}[scale=1]
		\begin{axis}[
		xlabel=Step,
		ylabel=Ratio of Safe Victims,
		mark options={scale=1},
				legend style ={ at={(0.98,0.02)}, 
					anchor=south east, draw=black, 
					fill=white,align=left},
		vasymptote=25
		]
		
		\addplot+ 
		table [x index=0, y index=1, y error index=2, col sep=space] {experiments/mctsGoalChange.txt};
		\addlegendentry{MCTS}
		\addplot+ 
		table [x index=0, y index=1, y error index=2, col sep=space] {experiments/mcapGoalChange.txt};
		\addlegendentry{MCAP}
		\end{axis}
		\end{tikzpicture}
		\label{fig:resultsEventsSafe}
	}
	\vfill
	\subfloat[]{
		\begin{tikzpicture}[scale=1]
		\begin{axis}[
		xlabel=Step,
		ylabel=Ratio of Burning Victims,
		mark options={scale=1},
		vasymptote=25
		]
		
		\addplot+ 
		table [x index=0, y index=3, y error index=4, col sep=space] {experiments/mctsGoalChange.txt};
		\addlegendentry{MCTS}
		\addplot+ 
		table [x index=0, y index=3, y error index=4, col sep=space] {experiments/mcapGoalChange.txt};
		\addlegendentry{MCAP}
		\end{axis}
		\end{tikzpicture}
		\label{fig:resultsEventsBurning}
	}
	\caption{Comparison of (a) safe victims and (b) burning ratios for MCTS and MCAP despite unexpected change of the reward function at step 25. See text for details.}
	\label{fig:resultsEvents}
\end{figure}
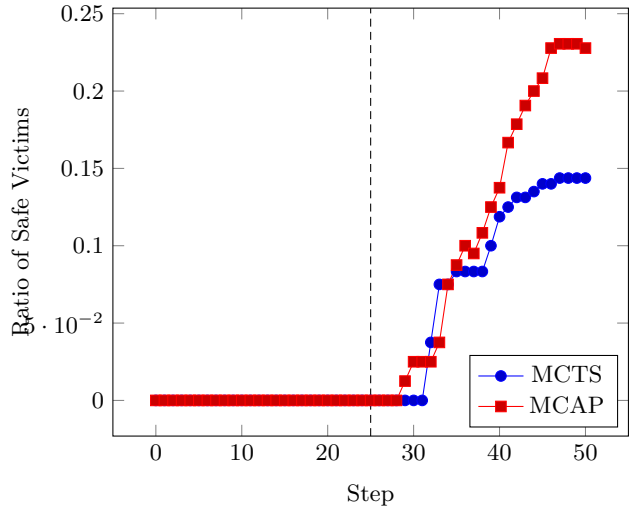
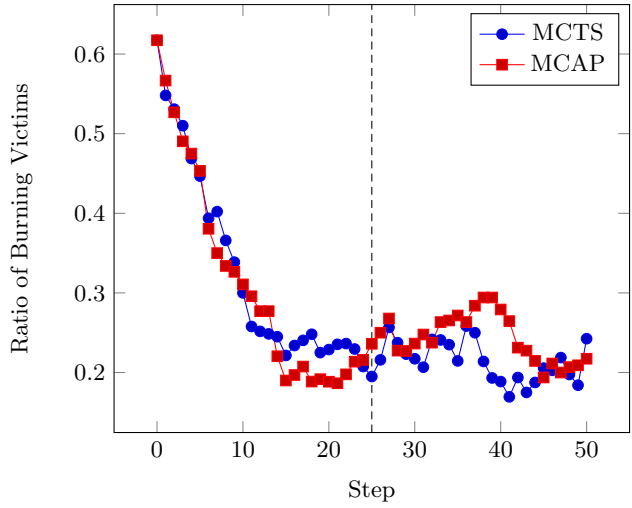

A similar experiment with unexpected events illustrates robustness of the approach. Here, every twenty steps all currently carried victims fell to the ground (i.e. were located at their carrier's position). Also, fires ignited such that overall at least ten fires were burning immediately after these events.
Note that the simulation of the domain used for plain MCTS and MCAP did \textit{not} simulate these events. The planning system managed to recover from the unexpected situations autonomously (Figure \ref{fig:resultsEvents}). As for the basic experiment, the configuration with MCAP performed significantly better that the configuration using plain MCTS.
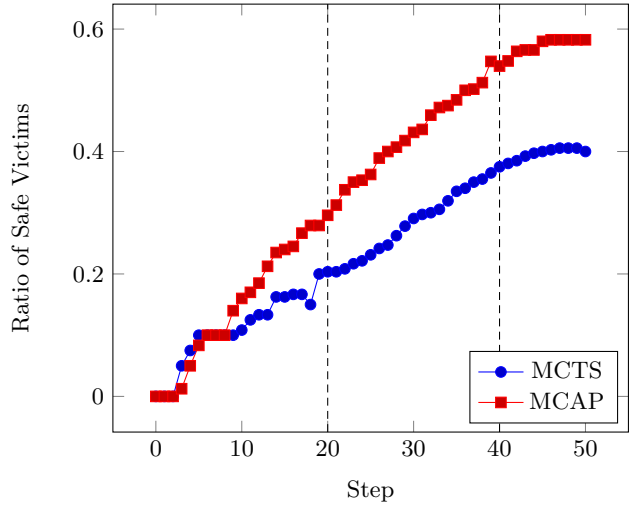
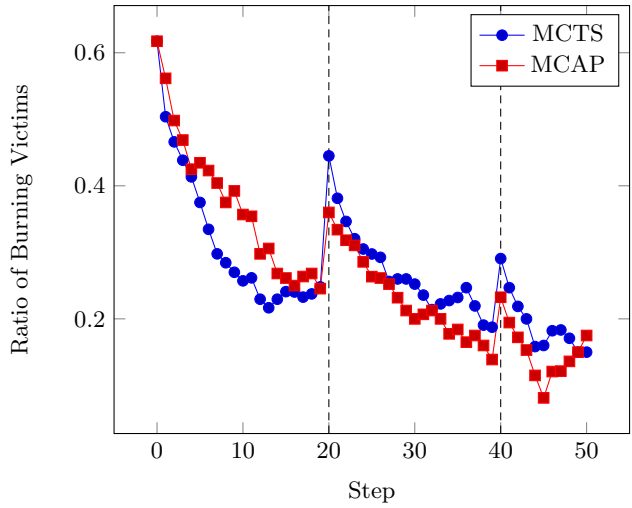
\begin{figure}
	\centering
	\subfloat[]{
		\begin{tikzpicture}[scale=1]
		\begin{axis}[
		xlabel=Step,
		ylabel=Ratio of Safe Victims,
		mark options={scale=1},
				legend style ={ at={(0.98,0.02)}, 
					anchor=south east, draw=black, 
					fill=white,align=left},
		vasymptote=20,
		vasymptote=40
		]
		
		\addplot+ 
		table [x index=0, y index=1, y error index=2, col sep=space] {experiments/mctsEvents.txt};
		\addlegendentry{MCTS}
		\addplot+ 
		table [x index=0, y index=1, y error index=2, col sep=space] {experiments/mcapEvents.txt};
		\addlegendentry{MCAP}
		\end{axis}
		\end{tikzpicture}
		\label{fig:resultsGoalChangeSafe}
	}
	\vfill
	\subfloat[]{
		\begin{tikzpicture}[scale=1]
		\begin{axis}[
		xlabel=Step,
		ylabel=Ratio of Burning Victims,
		mark options={scale=1},
		vasymptote=20,
		vasymptote=40
		]
		
		\addplot+ 
		table [x index=0, y index=3, y error index=4, col sep=space] {experiments/mctsEvents.txt};
		\addlegendentry{MCTS}
		\addplot+ 
		table [x index=0, y index=3, y error index=4, col sep=space] {experiments/mcapEvents.txt};
		\addlegendentry{MCAP}
		\end{axis}
		\end{tikzpicture}
		\label{fig:resultsGoalChangeBurning}
	}
	\caption{Comparison of (a) safe victims and (b) burning ratios for MCTS and MCAP despite unexpected events at steps 20 and 40. See text for details.}
	\label{fig:resultsGoalChange}
\end{figure}

In a third experiment the reward function was changed unexpectedly for the system. Before step 25, a reward is provided exclusively for avoiding burning victims. From step 25 on the reward function from the previous experiments was used, providing reward for safe victims. The planner did \textit{not} simulate the change of reward when evaluating action traces. MCAP outperformed plain MCTS by reacting more effectively to the change of reward function. Figure \ref{fig:resultsGoalChange} shows the results of this experiment.

\section{Conclusion}
\label{sec:conclusion}

This paper proposed Monte Carlo Action Programming, a programming language framework for autonomous systems that act in large probabilistic state spaces.
It comprises formal syntax and semantics of a nondeterministic action programming language. The language is interpreted stochastically via Monte Carlo Tree Search.
The effectiveness of search space constraint specification in the MCAP framework was shown empirically. Online interpretation of MCAP provides system performance and robustness in the face of unexpected events.

A possible venue for further research is the extension of MCAP to domains with continuous time and hybrid systems. Here, discrete programs are interpreted w.r.t. continuously evolving domain values \cite{alur1995algorithmic}.
%
%
It would also be interesting to evaluate to what extend manual specification techniques as MCAP could be combined with online representation learning (e.g. statistical relational learning \cite{getoor2007introduction} and deep learning \cite{hinton2006fast}): How to constrain system behavior if perceptual abstraction is unknown at design time or changes at runtime?


\bibliographystyle{splncs}
\bibliography{references}

\end{document}